\documentclass[11pt]{article}

\usepackage[final]{acl}
\usepackage{graphicx}
\usepackage{float}
\usepackage{booktabs} 
\usepackage{multirow} 
\PassOptionsToPackage{numbers,sort&compress}{natbib}
\usepackage{tabularx}
\usepackage{booktabs}
\usepackage{ragged2e}
\usepackage{hyperref}
\usepackage{times}
\usepackage{latexsym}
\usepackage{amsmath}
\usepackage{amsfonts} 
\usepackage{bm}
\newcommand{\sgraph}{S_{\mathrm{graph}}}

\newcommand{\scomm}{S_{\mathrm{comm}}}
\usepackage{enumitem}

\usepackage[T1]{fontenc}

\usepackage[utf8]{inputenc}

\usepackage{microtype}

\usepackage{inconsolata}

\usepackage{graphicx}

\title{OMD-GraphRAG: Enhancing GraphRAG with Ontology-Guided Extraction, Multi-Dimensional Clustering and Dual-Channel Fusion}

\author{
Jie Wang\textsuperscript{\textdagger}\textsuperscript{*} ,
Honghua Huang\textsuperscript{\textdagger} ,
Xi Ge\textsuperscript{\textdagger} ,
Jianhui Su\textsuperscript{\textdagger}, \\
Wen Liu\textsuperscript{\textdagger} ,
Shiguo Lian\textsuperscript{\textdagger}\textsuperscript{*} \\
\small $^\dagger$Equal contribution, $^*$Corresponding Authors \\
Data Science \& Artificial Intelligence Research Institute, China Unicom \\
\texttt{\{wangj1075, liansg, gex26, huanghh56\}@chinaunicom.cn} \\
}

\begin{document}
\maketitle
\begin{abstract}
Retrieval-Augmented Generation (RAG) systems face significant challenges in complex reasoning, multi-hop queries, and domain-specific QA. While existing GraphRAG frameworks have made progress in structural knowledge organization, they still have limitations in knowledge extraction precision, community report integrity, and retrieval performance. This paper proposes OMD-GraphRAG, an enhanced framework built upon open-source GraphRAG. The framework introduces three core innovations: (1) Ontology-Guided Knowledge Extraction that uses predefined Schema to guide LLMs in accurately identifying domain-specific entities and relations; (2) Multi-Dimensional Community Clustering Strategy that improves community completeness through alignment completion, attribute-based clustering, and multi-hop relationship clustering; (3) Dual-Channel Graph Retrieval Fusion that balances QA accuracy and performance through hybrid graph and community retrieval. Evaluation results on MultiHop-RAG benchmark show that OMD-GraphRAG outperforms mainstream open source solutions (e.g., LightRAG) in comprehensive F1 scores, particularly in inference and temporal queries.
\end{abstract}

\section{Introduction}

Large Language Models (LLMs) have demonstrated remarkable capabilities in natural language understanding and generation. However, they still suffer from hallucination and lack of domain-specific knowledge, particularly in vertical industries such as healthcare, finance, and legal domains. Retrieval-Augmented Generation (RAG) addresses these limitations by grounding LLM responses in external knowledge sources.

GraphRAG~\citep{7} extends traditional RAG by organizing fragmented knowledge into structured graphs, enabling complex multi-hop reasoning across documents~\citep{2,3,8}. Despite its promise, existing GraphRAG frameworks face three critical bottlenecks in vertical applications: (1)absence of extraction specification:schema-free extraction methods~\citep{11} yield loosely structured graphs,lacking predefined constraints on entities and relations, they result in uneven extraction quality that struggles to support complex reasoning chains;(2)single-dimensional community clustering:topology-only algorithms such as Leiden and Louvain~\citep{6} sever cross-community links and cannot aggregate nodes by business attributes such as time or location;  (3)inflexible single-channel retrieval:uniform retrieval cannot differentiate entity-level factoid queries from thematic multi-hop queries, and LLM-dependent online query routing~\citep{16} introduces per-query overhead that limits scalability.

To tackle these challenges, we present OMD-GraphRAG a framework that improves knowledge extraction precision via ontology-guided constraints, broadens the diversity of community themes by extending clustering dimensions, and enhances multi-hop retrieval performance through a dual-channel fusion of hybrid and graph retrieval. The three major differentiating advantages and contributions of OMD-GraphRAG are outlined below:

\begin{itemize}
    \item \textbf{Ontology-Guided Knowledge Extraction:} We propose a schema-guided extraction mechanism that injects predefined ontology templates into the LLM prompt and filters extracted triples via post-hoc type checking, aligning the knowledge graph with domain-specific entity hierarchies and reducing noise.Experiments demonstrate that this method improves retrieval accuracy by over \textbf{3.17\%} on the MultiHop-RAG dataset.

    \item \textbf{Multi-Dimensional Community Clustering:} We propose a clustering strategy that extends Leiden with boundary node completion, attribute-aware modularity optimization, and multi-hop relational subgraph construction, producing richer and more complete community reports.Experiments demonstrate that this method improves retrieval accuracy by over \textbf{3.43\%} on the MultiHop-RAG dataset.

    \item \textbf{Dual-Channel Graph Retrieval Fusion:} We propose a hybrid retrieval architecture that combines entity-level graph traversal with community-level semantic retrieval, adaptively weighted by query complexity, and re-ranked by a cross-encoder.Experiments demonstrate that this method improves retrieval accuracy by over \textbf{3.32\%} on the MultiHop-RAG dataset.
    
    \item Integrating above novel techniques on MultiHop-RAG~\citep{17}, Our OMD-GraphRAG improves average F1 by \textbf{9.21\%} over open-source GraphRAG implementation(LightRAG)~\citep{4},demonstrating the effectiveness of our method.
\end{itemize}

\begin{figure*}[h]
  \centering
  \includegraphics[width=0.9\linewidth]{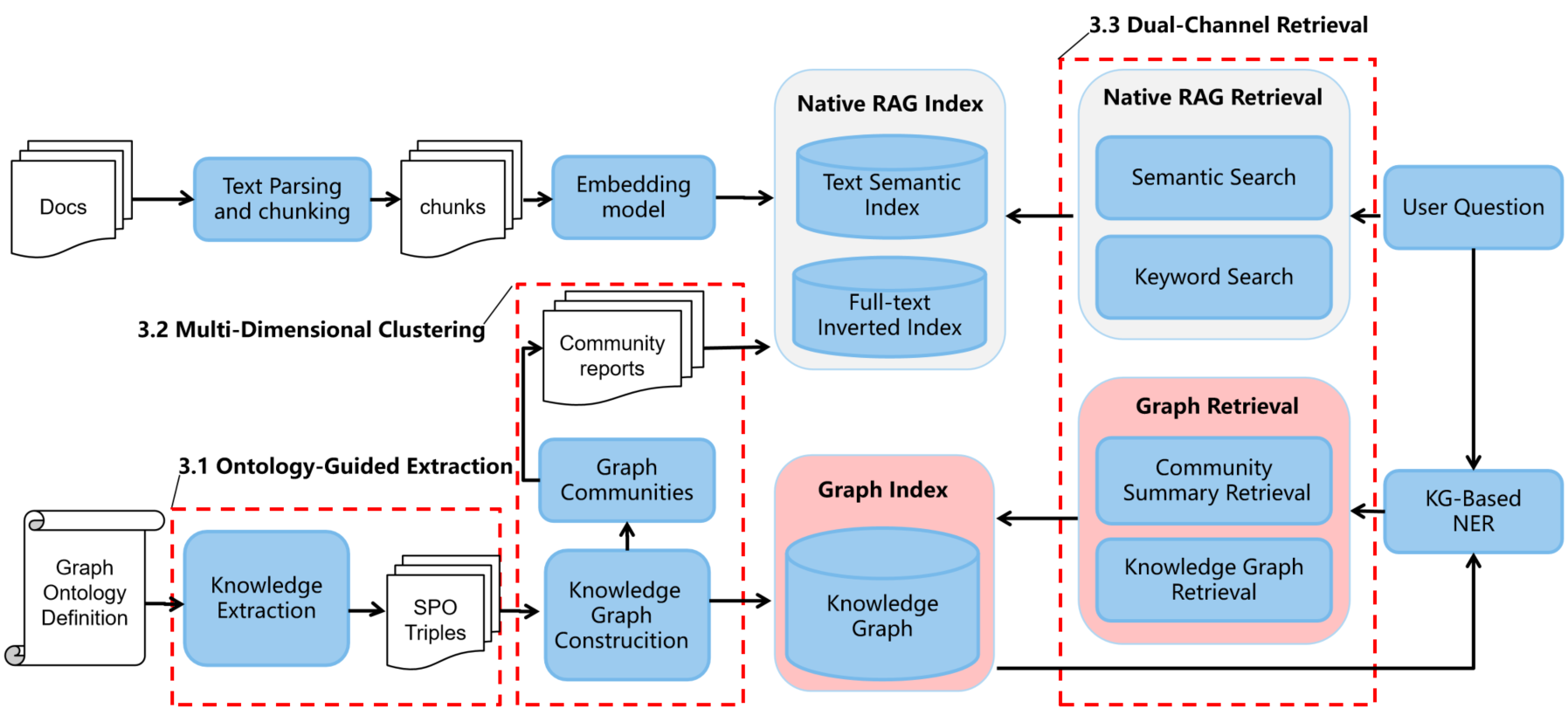}
  \caption{Architecture of OMD-GraphRAG: Enhancing with Ontology-Guided Extraction,
Multi-Dimensional Clustering and Dual-Channel Fusion}
  \label{fig:design}
\end{figure*}  

\section{Related Work}

GraphRAG extends foundational RAG~\citep{9,10} by organizing document corpora into structured knowledge graphs to support multi-hop reasoning and global summarization~\citep{7}, with pipelines spanning knowledge graph extraction, community detection, and graph-enhanced retrieval~\citep{2,3,8}.

\paragraph{Knowledge Graph Extraction.}
GraphRAG systems~\citep{7,4} extract SPO triples via unconstrained prompting sacrificing schema coherence for indexing speed. Youtu-GraphRAG~\citep{dong2026youtu} uses a seed graph schema to bound extraction to targeted types, achieving robust domain transfer. Bu et al.~\citep{bu2025querydriven} derive query-driven graph patterns dynamically, but rely on query semantics rather than a fixed ontology, limiting schema consistency. Prior LLM-based KG extraction~\citep{5,11} suffers from inconsistent typing and fragmented relations due to schema-free extraction, a limitation our ontology-guided approach addresses.

\paragraph{Community Detection and Organization.}
GraphRAG~\citep{7} applies hierarchical Leiden clustering~\citep{6} to build multi-level community summaries for query-focused summarization. HyperGraphRAG~\citep{luo2025hypergraphrag} replaces binary edges with hyperedges to model $n$-ary relations, while HELP~\citep{huang2026help} gains efficiency through precomputed graph-text correlations. Despite these advances, non-overlapping clustering severs inter-community edges, purely topological modularity ignores node attributes, and intra-community subgraphs rarely preserve complete multi-hop inference paths. Our Multi-Dimensional Community Clustering targets all three gaps through $\epsilon$-neighbor completion, attribute-aware modularity, and path-pattern-constrained subgraph construction.

\paragraph{Graph-Enhanced Retrieval.}
LightRAG~\citep{4} introduces dual local-global retrieval switching between entity-level and community-level contexts. GraphSearch~\citep{yang2025graphsearch} integrates semantic and structural signals through multi-turn agentic reasoning, while Wang et al.~\citep{wang2026hypergraph} fuse entity and passage similarities via hypergraph diffusion. These approaches either incur online latency through retrieval-time traversal or fail to capture complementary evidence across granularities. Our static dual-channel fusion addresses both by encoding complementary evidence at indexing time.

\section{OMD-GraphRAG}

Our OMD-GraphRAG follows the Index-Retrieve-Generate paradigm with specific optimizations as shown in Figure \ref{fig:design}:

\begin{itemize}
    \item \textbf{Indexing:} Quality domain graph construction via ontology-guided extraction.

    \item \textbf{Retrieval:} A dual-channel fusion mode that dynamically selects the optimal path.

    \item \textbf{Generation:} Utilizing enhanced community reports and entity context for comprehensive answers.
\end{itemize}

\subsection{Ontology-Guided Knowledge Extraction}

\begin{figure*}[h]
  \centering
  \includegraphics[width=0.9\linewidth]{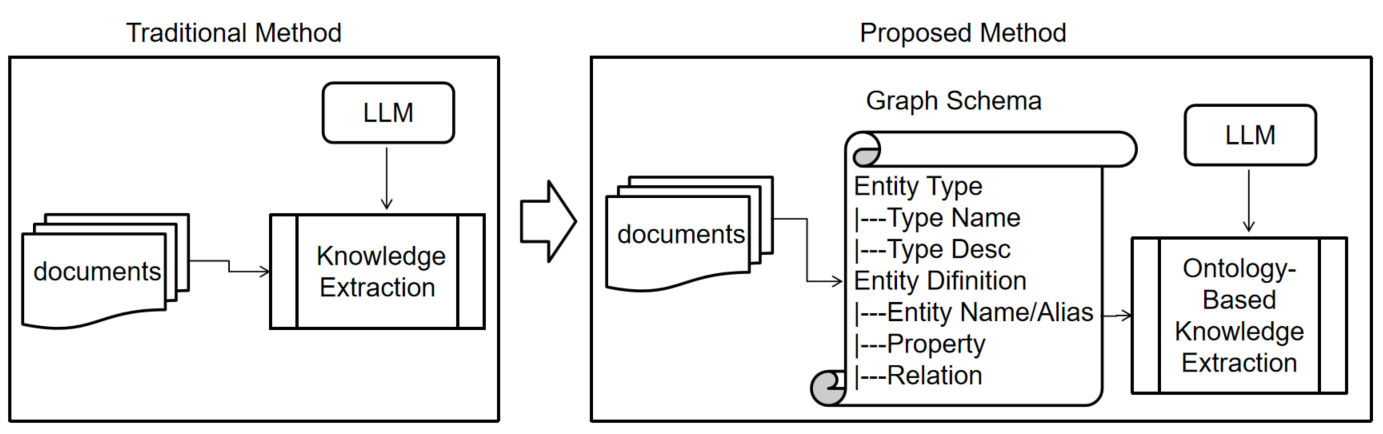}
  \caption{Proposing an innovative approach to knowledge extraction guided by ontology constraints, building upon traditional LLM methods.}
  \label{fig:entity}
\end{figure*}  

As shown in Figure \ref{fig:entity}, the framework supports domain expert-defined Schema templates. By inputting these Schemas as prompts, the LLM is guided to extract SPO triples with high accuracy, effectively reducing noise and improving cross-domain migration.
In the construction of knowledge graphs, an ontology ($\mathcal{O}$) is defined as a set of logical constraints governing domain concepts and their interrelations. We formalize the Ontology Schema as a triplet constraint space $\mathcal{S} = (\mathcal{E}, \mathcal{R}, \Phi)$, where:
\begin{itemize}
    \item $\mathcal{E}$ denotes the set of permissible entity types. For example, in a cultural heritage domain: $\mathcal{E} = \{\texttt{Artifact},\,\texttt{Museum}\}$.
    \item $\mathcal{R}$ represents the set of allowable relation types. For example: $\mathcal{R} = \{\texttt{CollectedBy}\}$.
    \item $\Phi$ serves as the type constraint function, defining the valid domain and range for each relation:
    \begin{equation}
        \Phi(r) = (\text{dom}(r), \text{range}(r)), \quad \forall r \in \mathcal{R}
    \end{equation}
    For the example above, $\Phi(\texttt{CollectedBy}) = (\texttt{Artifact},\,\texttt{Museum})$: only triples of the form $\langle\textit{artifact},\,\texttt{CollectedBy},\,\textit{museum}\rangle$ pass type checking; triples with mismatched head or tail types are discarded post-hoc.
\end{itemize}
In contrast to conventional open information extraction, which maximizes $P(T|D)$ without structural constraints, OMD-GraphRAG adopts a \textit{schema-guided extraction} approach: the ontology schema $\mathcal{S}$ is injected into the LLM prompt as a structured instruction, and each generated triple $(h, r, t_{\text{tail}})$ is subsequently validated by checking $\text{type}(h) \in \text{dom}(r)$ and $\text{type}(t_{\text{tail}}) \in \text{range}(r)$ against $\Phi$; triples that violate these constraints are discarded post-hoc.

\subsection{Multi-Dimensional Community Clustering Strategy}

\begin{figure*}[h]
  \centering
  \includegraphics[width=1.0\linewidth]{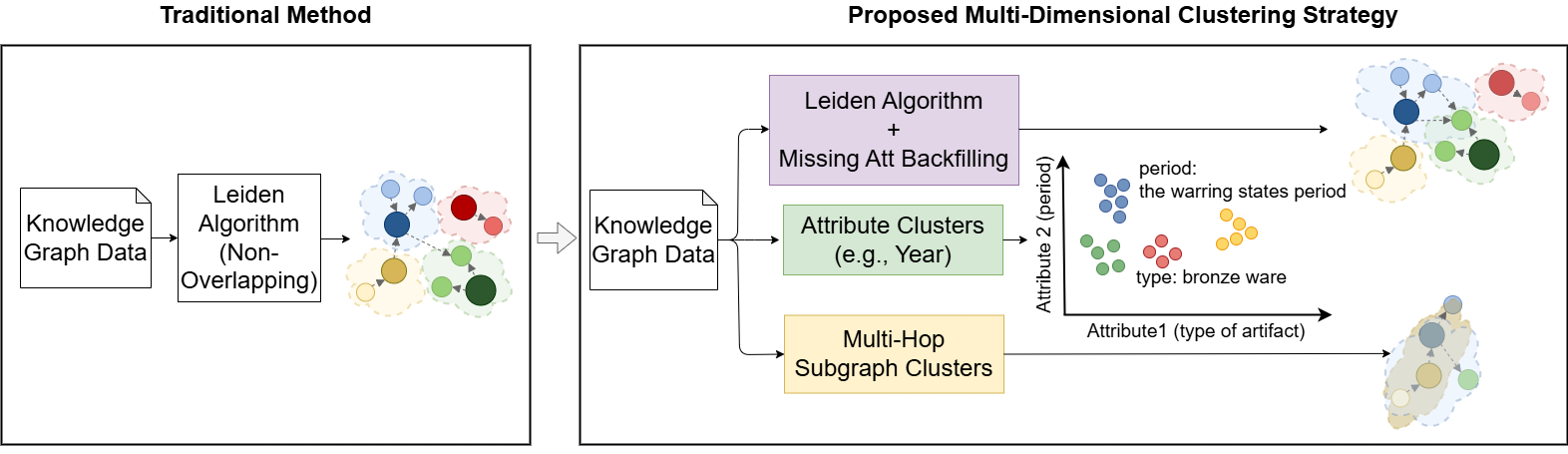}
  \caption{Multi-Dimensional Community Clustering. We propose three clustering approaches built upon the Leiden algorithm.}
  \label{fig:community}
\end{figure*}

As shown in Figure \ref{fig:community}, we introduce three key optimizations to improve the quality of community reports:

\begin{itemize}
    \item \textbf{Attribute-Based Clustering:} Allows clustering by dimensions such as "year" or "location," which is highly effective for queries like "Which artifacts in the museum are from the Warring States period?".

    \item \textbf{Alignment \& Completion Mechanism:} Automatically backfills missing attributes or relationships for leaf nodes that were severed by the non-overlapping nature of the Leiden algorithm.

    \item \textbf{Multi-Hop Relationship Clustering:} Constructs subgraphs via N-hop path-pattern-constrained expansion, clustering nodes reachable within $N$ hops along specific relational patterns (e.g., \texttt{Cause $\to$ Effect}), directly serving multi-hop QA tasks.

\end{itemize}

\subsubsection{Improved Modularity Function with Attribute Awareness}

Conventional Louvain and Leiden algorithms optimize solely for topological modularity, denoted as $Q$. To incorporate node attributes into the community detection process, we propose the Attribute-Aware Modularity metric, defined as:

\begin{center}
\small
\begin{equation}
    Q_{\text{multi}} = \frac{1}{2m} \sum_{i,j} \left( Q_{ij}^{\text{struct}} + \alpha \cdot S_{ij} \right) \delta(c_i, c_j)
    \label{eq:compact_modularity}
\end{equation}
\normalsize
\end{center}

\noindent where:
\begin{itemize}
    \item $Q_{ij}^{\text{struct}} = A_{ij} - \frac{k_i k_j}{2m}$ represents the structural contribution, measuring whether nodes $i$ and $j$ are more connected than expected by chance.
    \item $S_{ij} = \text{Sim}(a_i, a_j)$ denotes the attribute similarity, quantifying the semantic closeness between the features of nodes $i$ and $j$.
    \item $\alpha$ is a balancing factor that adjusts the relative importance of attribute similarity against the network structure.
    \item $\delta(c_i, c_j)$ is the Kronecker delta function, which equals $1$ if nodes $i$ and $j$ belong to the same community, and $0$ otherwise.
\end{itemize}

This formulation provides the theoretical foundation for why clustering based on attributes such as ``artifact dynasty'' yields communities with tighter semantic cohesion.

\subsubsection{Alignment \& Completion Mechanism for Community Boundary}
\label{sec:completion}

To address the issue of \textit{inter-community edge severance} inherent in non-overlapping clustering, we introduce an $\epsilon$-neighbor completion mechanism. For any node $v$ within a community $C_k$, if a neighbor $u \notin C_k$ exhibits a connection strength to nodes within $C_k$ exceeding a threshold $\tau$, the attributes and relations of $u$ are \textit{projected} into the summary representation of $C_k$. Formally, the completed community is defined as:
\begin{center}
\small
\begin{equation}
    \widehat{C}_k = C_k \cup \left\{ u \mid \exists v \in C_k, (v, u) \in E \land \frac{\sum_{w \in C_k} A_{uw}}{\deg(u)} \geq \tau \right\}
    \label{eq:boundary_completion}
\end{equation}
\normalsize
\end{center}
\noindent where:
\begin{itemize}
    \item $\widehat{C}_k$ denotes the completed community, extending the original set $C_k$ with relevant boundary nodes.
    \item $A_{uw}$ is the adjacency indicator, equal to $1$ if an edge exists between the external node $u$ and an internal node $w$, and $0$ otherwise.
    \item $\sum_{w \in C_k} A_{uw}$ counts the internal degree of node $u$, representing the number of connections $u$ has specifically within community $C_k$.
    \item $\deg(u)$ is the total degree of node $u$ in the entire graph, serving as a normalization factor.
    \item The ratio $\frac{\sum_{w \in C_k} A_{uw}}{\deg(u)}$ represents the affinity score, quantifying the proportion of $u$'s total connections that point towards $C_k$.
    \item $\tau \in [0, 1]$ is the inclusion threshold; only neighbors with an affinity score above $\tau$ are absorbed into the community.
\end{itemize}
\noindent This mechanism ensures the \textbf{logical completeness} of community reports and prevents the fragmentation of critical \textit{reasoning chains}.

\subsubsection{Path-Pattern-Constrained Multi-Hop Subgraph Construction}

For multi-hop relational clustering, we construct a subgraph rooted at a seed node $v_{root}$ by including all nodes reachable within $N$ hops whose connecting path matches a predefined relational pattern (e.g., \texttt{Cause $\to$ Effect}). This constraint ensures that clustered nodes form semantically coherent inference chains rather than arbitrary topological neighborhoods, directly serving Multi-Hop Question Answering (QA) tasks.

\subsection{Dual-Channel Graph Retrieval Fusion Mode}

\begin{figure*}[h]
  \centering
  \includegraphics[width=0.9\linewidth]{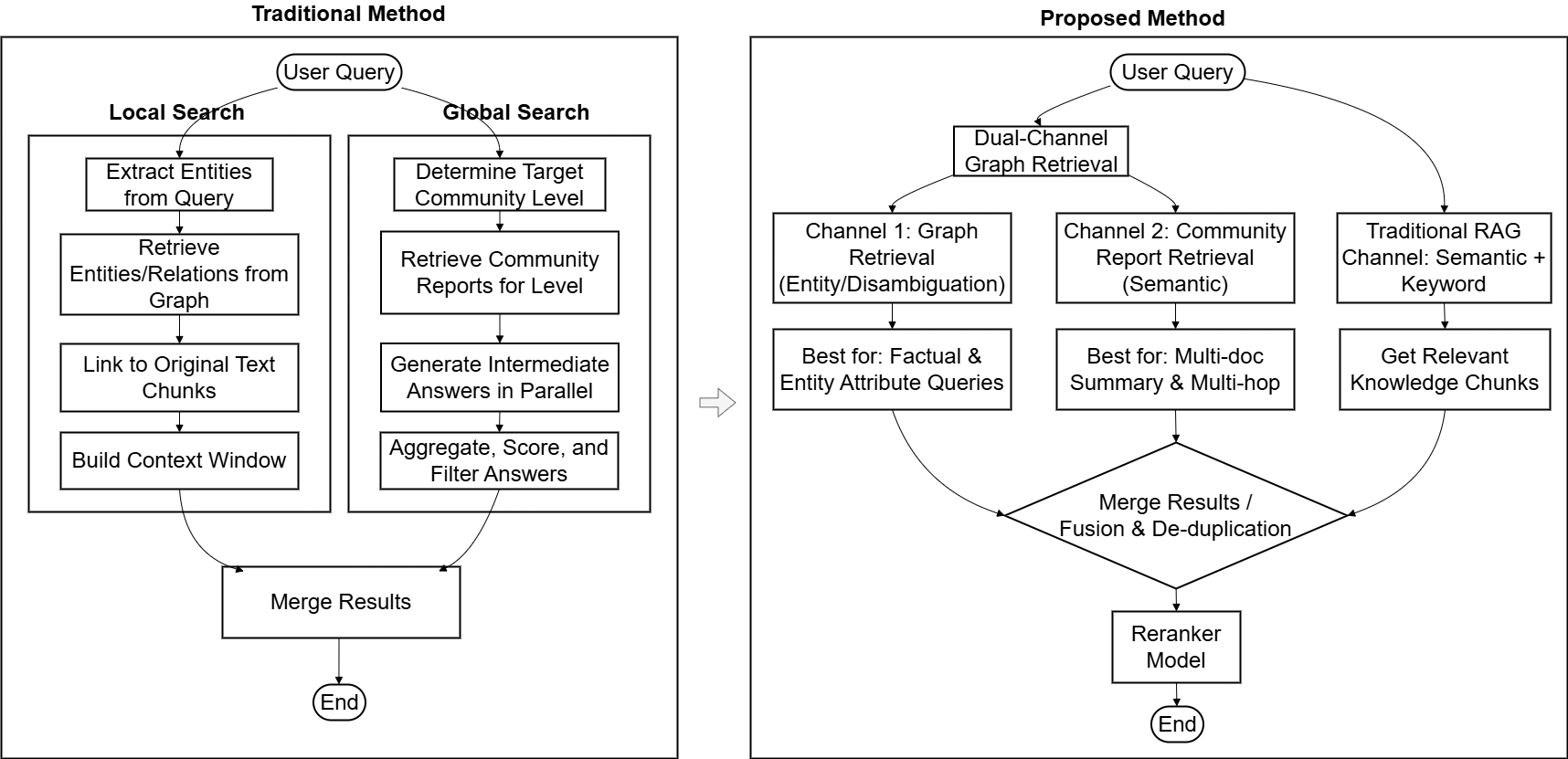}
  \caption{Innovating a dual-channel retrieval mechanism that combines SPO and community report retrieval on top of traditional RAG hybrid search, achieving a balance between retrieval precision and performance.}
  \label{fig:retrieval}
\end{figure*}  

As shown in Figure \ref{fig:retrieval}, rather than relying solely on expensive LLM-based intent classification, this framework uses two optimized channels:

\begin{itemize}
    \item \textbf{Channel 1 (Knowledge Graph Retrieval):} Uses trie\footnote{A trie (prefix tree) is a data structure enabling $O(|q|)$ prefix-matched entity lookup.} matching for dynamic entity disambiguation and attribute traversal. Best for factual and entity-attribute queries.

    \item \textbf{Channel 2 (Community Report Retrieval):} Matches query semantic themes against multi-dimensional community topics. Best for multi-document summarization~\citep{18} and complex relationship queries. The results are fused with traditional RAG results and passed through a Reranker model for final output.
\end{itemize}

\subsubsection{Fusion Model for Dual-Channel Retrieval}
Let $q$ denote the user query.
\begin{itemize}
    \item \textbf{Channel I (Entity-Level):} Entities in the query are matched via trie lookup; the $k$-hop neighborhood subgraph $\mathcal{N}(e)$ of each matched entity $e$ is traversed, and candidate document chunks are scored by their structural overlap with $\mathcal{N}(e)$. The channel score $\sgraph(d_i | q)$ aggregates these overlap scores across all query entities.

    \item \textbf{Channel II (Community-Level):} The retrieval score $\scomm$ is based on the semantic vector similarity between the query and community summaries:
    \begin{equation}
        \scomm(C_k | q) = \cos(\bm{v}_q, \bm{v}_{C_k}) = \frac{\bm{v}_q \cdot \bm{v}_{C_k}}{\|\bm{v}_q\| \|\bm{v}_{C_k}\|}
    \end{equation}
    where $\bm{v}_{C_k}$ is the embedding vector of the multi-dimensional community report $C_k$.
\end{itemize}
\subsubsection{Dynamic Weighted Fusion Strategy}
To adaptively integrate these two retrieval signals, we introduce a query complexity-aware weight $\beta(q)$. For \textit{factoid queries} (typically characterized by high entity density), $\beta$ biases towards the graph channel to ensure factual precision; conversely, for \textit{summarization or multi-hop queries} (characterized by high semantic abstraction), $\beta$ favors the community channel to capture global themes. The final retrieval score $\mathcal{S}_{\text{final}}$ for a document $d_i$ given query $q$ is defined as:

\begin{equation}
\begin{split}
    \mathcal{S}_{\text{final}}(d_i | q) = &\;\beta(q) \cdot \mathcal{S}_{\text{graph}}(d_i | q) \\
    &+ (1 - \beta(q)) \cdot \max_{C_k: d_i \in C_k} \mathcal{S}_{\text{comm}}(C_k | q)
\end{split}
    \label{eq:dynamic_fusion}
\end{equation}

The adaptive weight $\beta(q) \in (0, 1)$ is computed via a lightweight heuristic that balances explicit entity presence against semantic abstraction:
\begin{equation}
    \beta(q) = \sigma \left( w_1 \cdot \tilde{N}_{\text{ent}}(q) - w_2 \cdot H_{\text{sem}}(q) \right)
    \label{eq:beta_calc}
\end{equation}
\noindent where:
\begin{itemize}
    \item $\sigma$ is the Sigmoid function mapping the score to $(0, 1)$.
    \item $\tilde{N}_{\text{ent}}(q)$ is the normalized entity density, ensuring scale invariance across query lengths.
    \item $H_{\text{sem}}(q)$ denotes the semantic abstraction score (non-entity token entropy); higher values indicate topic-oriented queries requiring global reasoning.
    \item $\max_{C_k: d_i \in C_k}$ selects the highest relevance among all communities containing $d_i$, ensuring the dominant theme drives the fusion.
\end{itemize}
This strategy automatically routes fact-based queries to graph traversal and insight-based queries to community reports.

\subsubsection{Re-ranking Phase}
In the final stage, a cross-encoder reranker (\texttt{qwen3-reranker-8b}) scores each candidate in the fused set $\mathcal{D}_{cand}$ against query $q$ and selects the top-$K$ results. By supplying the reranker with candidates from both the graph and community channels, we increase diversity in the candidate pool, providing complementary micro-level facts and macro-level context for final answer generation. 

\section{Experiments and Evaluation}

\subsection{Dataset}

We conducted a comprehensive evaluation on the \textit{MultiHop-RAG} benchmark \citep{17}, an English-language dataset accepted at COLM 2024 specifically designed to assess multi-hop reasoning in RAG systems. The underlying data source comprises various English-language websites covering a wide range of news categories, including entertainment, business, sports, technology, health, and science. It contains 2{,}556 queries whose evidence spans 2 to 4 documents each, and incorporates document metadata to reflect complex real-world RAG scenarios. The benchmark focuses on evaluating logical reasoning across multiple documents, temporal dependencies, and attribute comparisons.

The dataset includes four distinct categories of queries:inference query,comparison query,and temporal query,we selected the following three categories as our evaluation set:
\begin{itemize}
    \item \textbf{Inference Query:} Requires logical derivation from multiple scattered facts.
    \item \textbf{Comparison Query:} Involves horizontal comparison of event attributes.
    \item \textbf{Temporal Query:} Requires understanding the chronological order and time dependencies of events.
\end{itemize}

Excluding Null Query:For null queries, the dataset authors introduced this category specifically to evaluate generation quality, particularly the model's ability to refuse answering \citep{17}. For such queries, standard retrieval metrics cannot be derived based on ground truth; the expected ground truth for retrieval is an empty set. However, under traditional RAG retrieval paradigms, a given query will invariably return some results from the document corpus regardless of relevance. Consequently, the Null Query category cannot effectively assess the Relevancy and Recall of retrieval performance. As this category lacks reference value for evaluating retrieval capabilities, it has been excluded from our evaluation scope.

Detailed descriptions of the query types are provided in Appendix~\ref{app:query_types}.

\subsection{Metrics}

Following the evaluation protocol of the MultiHop-RAG benchmark~\citep{17}, all metrics are computed by an LLM judge over (query, retrieved context, generated answer) tuples and reported as percentages:

\begin{itemize}
    \item \textbf{Relevancy:}\footnote{We adopt the metric name ``Relevancy'' as defined in the original MultiHop-RAG benchmark \citep{17}.} The degree to which the generated answer addresses the query, judged on a continuous scale.

    \item \textbf{Recall:} The proportion of key factual content from the ground-truth answer that is covered by the generated response.

    \item \textbf{F1 Score:} The harmonic mean of Relevancy and Recall: $\text{F1} = \frac{2 \times \text{Relevancy} \times \text{Recall}}{\text{Relevancy} + \text{Recall}}$.
\end{itemize}

\subsection{Baseline Methods}

We compared OMD-GraphRAG against the following open-source frameworks and internal baselines:

\begin{itemize}[nosep]
    \item \textbf{Dify Naive RAG:} A traditional RAG system combining dense vector retrieval with BM25-based keyword retrieval.
    \item \textbf{Open-source LightRAG:} \citep{4} A lightweight GraphRAG system representing efficient graph-based retrieval architectures.
\end{itemize}

Additional implementation details are also provided in Appendix~\ref{app:implementation}.


\subsection{Results Analysis}

\begin{table}[H]
    \centering
    \caption{Performance Comparison on MultiHop-RAG (best scores are in bold, second-best results are underlined)}
    \label{tab:results_comparison}
    \setlength{\tabcolsep}{2.5pt}
    \resizebox{\columnwidth}{!}{
        \begin{tabular}{@{}llccc@{}}
            \toprule
            \textbf{Query Type} & \textbf{Method} & \textbf{Rel. (\%)} & \textbf{Rec. (\%)} & \textbf{F1} \\ \midrule
            
            \multirow{4}{*}{Inference} 
            & NaiveRAG(Dify)      & 67.86 & 56.71 & 61.79 \\
            & GraphRAG(LightRAG) & \underline{94.14} & \underline{83.16} & \underline{88.31} \\
            & \textbf{OMD-GraphRAG} & \textbf{98.35} & \textbf{83.86} & \textbf{90.53} \\ \midrule

            \multirow{4}{*}{Comparison} 
            & NaiveRAG(Dify)      & 43.78 & 54.75 & 48.65 \\
            & GraphRAG(LightRAG) & \underline{52.97} & 80.26 & \underline{63.82} \\
            & \textbf{OMD-GraphRAG} & \textbf{74.69} & \underline{81.62} & \textbf{78.97} \\ \midrule

            \multirow{4}{*}{Temporal} 
            & NaiveRAG(Dify)      & 26.37 & 53.00 & 35.22 \\
            & GraphRAG(LightRAG) & \underline{36.46} & \textbf{79.51} & \underline{49.99} \\
            & \textbf{OMD-GraphRAG} & \textbf{56.37} &  \underline{77.12} & \textbf{65.13} \\ \midrule

            \multirow{4}{*}{\textbf{Average}} 
            & NaiveRAG(Dify)      & 46.00 & 54.82 & 50.03 \\
            & GraphRAG(LightRAG) & \underline{61.19} & \textbf{80.98} & \underline{69.71} \\
            & \textbf{OMD-GraphRAG} & \textbf{77.07} & \underline{80.87} & \textbf{78.92} \\ \bottomrule
        \end{tabular}
    }
\end{table}

OMD-GraphRAG completed the performance benchmarking on the MultiHop-RAG evaluation set. The comparative results of performance metrics against mainstream methods across three distinct query categories are summarized in Table~\ref{tab:results_comparison}.

The core findings of this study can be summarized as follows:

\begin{itemize}[nosep]
\item \textbf{Consistent Superiority:} OMD-GraphRAG achieves the optimal F1-score across all three query categories, demonstrating consistent improvement in diverse multi-hop scenarios. We note that results reflect single-run evaluations; statistical significance testing is left for future work.

\item \textbf{Quantifiable Gains:} Compared to Dify Naive RAG, the overall F1-score is improved by \textbf{28.89\%}, and it outperforms the state-of-the-art Open-LightRAG by \textbf{9.21\%}.

\item \textbf{Relational Reasoning:} The marked advantage in Temporal Queries validates the effectiveness of our multi-hop relationship clustering strategy in resolving complex time-dependent dependencies.

\item \textbf{Query-Type Analysis:} Comparison queries require cross-document attribute alignment, reflecting an inherent retrieval difficulty (best F1: 78.97\% across all systems). Nonetheless, OMD-GraphRAG improves over LightRAG by \textbf{15.15\%} on this type, suggesting that ontology-guided extraction is particularly effective for identifying and aligning comparable entities.
\end{itemize}

\subsection{Ablation Experiment}
\subsubsection{Impact of Ontology-Guided Knowledge Extraction}

To evaluate the contribution of the ontology-guided extraction module, we compared a standard Naive RAG baseline (Vector + Keyword search) against an enhanced variant integrating our proposed ontology-guided knowledge extraction with triplet-based retrieval. All experiments were conducted on the MultiHop-RAG dataset across three query dimensions: Inference, Comparison, and Temporal.With retrieval $top\text{-}k$ set to 8, using \texttt{qwen3-embed-0.6b} for embeddings and \texttt{qwen3-reranker-8b} for reranking. Performance was measured via \textit{Relevancy}, \textit{Recall}, and \textit{F1-score}. Results are presented in Table~\ref{tab:ablation_results}.

\begin{table}[H]
    \centering
    \caption{Ablation Study Results on MultiHop-RAG}
    \label{tab:ablation_results}
    \setlength{\tabcolsep}{3pt} 
    \small
    \resizebox{\columnwidth}{!}{
        \begin{tabular}{@{}llccc@{}}
            \toprule
            \textbf{Query Type} & \textbf{Method} & \textbf{Relevancy (\%)} & \textbf{Recall (\%)} & \textbf{F1 (\%)} \\ \midrule
            
            \multirow{2}{*}{Inference} 
            & \textbf{w/ Ontology Extraction} & \textbf{97.51} & \textbf{83.50} & \textbf{89.96} \\
            & Naive RAG & 97.06 & 82.99 & 89.48 \\ \midrule

            \multirow{2}{*}{Comparison} 
            & \textbf{w/ Ontology Extraction} & \textbf{74.02} & 81.92 & \textbf{77.77} \\
            & Naive RAG & 67.93 & \textbf{82.78} & 74.62 \\ \midrule

            \multirow{2}{*}{Temporal} 
            & \textbf{w/ Ontology Extraction} & \textbf{56.68} & \textbf{79.60} & \textbf{66.21} \\
            & Naive RAG & 46.30 & 79.02 & 58.39 \\ \midrule

            \multirow{2}{*}{\textbf{Average}} 
            & \textbf{w/ Ontology Extraction} & \textbf{76.07} & \textbf{81.67} & \textbf{78.77} \\
            & Naive RAG & 70.43 & 81.60 & 75.60 \\ \bottomrule
        \end{tabular}
    }
\end{table}

By integrating ontology-guided knowledge extraction (the \textit{w/ Ontology Extraction} configuration) into the Naive RAG baseline, the integrated F1-score on the MultiHop-RAG dataset increased by \textbf{3.17\%} (from 75.60\% to 78.77\%). This improvement empirically validates the critical importance of ontology-based schema guidance in enhancing knowledge retrieval precision, particularly for addressing semantic gaps in unstructured data.

\subsubsection{Multi-dimensional Community Clustering Strategy}

To evaluate the effectiveness of our multi-hop relationship clustering and community summary retrieval, we compared a standard Naive RAG baseline (Vector + Keyword search) against an enhanced variant integrating our proposed multi-dimensional community report retrieval and recall mechanism. All experiments were conducted on the MultiHop-RAG benchmark across three query types: Inference, Comparison, and Temporal.With retrieval $top\text{-}k$ set to 8, using \texttt{qwen3-embed-0.6b} for embeddings and \texttt{qwen3-reranker-8b} for reranking. The boundary inclusion threshold $\tau$ introduced in Section~\ref{sec:completion} is set to $0.5$ in all experiments. Performance was assessed via \textit{Relevancy}, \textit{Recall}, and \textit{F1-score}. Results are detailed in Table~\ref{tab:ablation_community}.

\begin{table}[H]
    \centering
    \caption{Ablation Study Results: Multi-dimensional Community Strategy}
    \label{tab:ablation_community}
    \setlength{\tabcolsep}{3pt}
    \small
    \resizebox{\columnwidth}{!}{
        \begin{tabular}{@{}llccc@{}}
            \toprule
            \textbf{Query Type} & \textbf{Method} & \textbf{Relevancy (\%)} & \textbf{Recall (\%)} & \textbf{F1 (\%)} \\ \midrule
            
            \multirow{2}{*}{Inference} 
            & \textbf{w/ Multi-dim Community} & \textbf{98.07} & \textbf{84.30} & \textbf{90.67} \\
            & Naive RAG & 97.06 & 82.99 & 89.48 \\ \midrule

            \multirow{2}{*}{Comparison} 
            & \textbf{w/ Multi-dim Community} & \textbf{76.73} & 81.62 & \textbf{79.10} \\
            & Naive RAG & 67.93 & \textbf{82.78} & 74.62 \\ \midrule

            \multirow{2}{*}{Temporal} 
            & \textbf{w/ Multi-dim Community} & \textbf{57.29} & 76.39 & \textbf{65.48} \\
            & Naive RAG & 46.30 & \textbf{79.02} & 58.39 \\ \midrule

            \multirow{2}{*}{\textbf{Average}} 
            & \textbf{w/ Multi-dim Community} & \textbf{77.36} & 80.77 & \textbf{79.03} \\
            & Naive RAG & 70.43 & \textbf{81.60} & 75.60 \\ \bottomrule
        \end{tabular}
    }
\end{table}

Incorporating the multi-dimensional community report strategy (the \textit{w/ Multi-dim Community} configuration) into the Naive RAG baseline resulted in a \textbf{3.43\%} increase in the integrated F1-score (from 75.60\% to 79.03\%) on the MultiHop-RAG dataset. This improvement underscores the effectiveness of our multi-dimensional community clustering strategy in optimizing retrieval performance, as it effectively captures higher-level semantic structures and global information across the knowledge graph.

\subsubsection{Dual-Channel Graph Retrieval Fusion Mode}

To evaluate the effectiveness of the hybrid graph-based retrieval strategy, we compared a standard \textit{Naive RAG} baseline (Vector + Keyword search) against an enhanced variant integrating our proposed dual-channel graph retrieval fusion strategy, which combines fine-grained Knowledge Graph triplet retrieval with coarse-grained Community Report recall. All experiments were conducted on the MultiHop-RAG benchmark across three query types:Inference, Comparison, and Temporal.With retrieval $top\text{-}k$ set to 8, using \texttt{qwen3-embed-0.6b} for embeddings and \texttt{qwen3-reranker-8b} for reranking. Performance was assessed via \textit{Relevancy}, \textit{Recall}, and \textit{F1-score}. Results are summarized in Table~\ref{tab:ablation_dual_channel}.

\begin{table}[H]
    \centering
    \caption{Ablation Study Results: Dual-Channel Retrieval Fusion}
    \label{tab:ablation_dual_channel}
    \setlength{\tabcolsep}{3pt}
    \small
    \resizebox{\columnwidth}{!}{
        \begin{tabular}{@{}llccc@{}}
            \toprule
            \textbf{Query Type} & \textbf{Method} & \textbf{Relevancy (\%)} & \textbf{Recall (\%)} & \textbf{F1 (\%)} \\ \midrule
            
            \multirow{2}{*}{Inference} 
            & \textbf{w/ Dual-channel Retrieval} & \textbf{98.35} & \textbf{83.86} & \textbf{90.53} \\
            & Naive RAG & 97.06 & 82.99 & 89.48 \\ \midrule

            \multirow{2}{*}{Comparison} 
            & \textbf{w/ Dual-channel Retrieval} & \textbf{76.49} & 81.62 & \textbf{78.97} \\
            & Naive RAG & 67.93 & \textbf{82.78} & 74.62 \\ \midrule

            \multirow{2}{*}{Temporal} 
            & \textbf{w/ Dual-channel Retrieval} & \textbf{56.37} & 77.12 & \textbf{65.13} \\
            & Naive RAG & 46.30 & \textbf{79.02} & 58.39 \\ \midrule

            \multirow{2}{*}{\textbf{Average}} 
            & \textbf{w/ Dual-channel Retrieval} & \textbf{77.07} & 80.87 & \textbf{78.92} \\
            & Naive RAG & 70.43 & \textbf{81.60} & 75.60 \\ \bottomrule
        \end{tabular}
    }
\end{table}

Incorporating the dual-channel graph retrieval fusion strategy (the \textit{w/ Dual-channel Retrieval} configuration) into the Naive RAG baseline resulted in a \textbf{3.32\%} improvement in the integrated F1-score (from 75.60\% to 78.92\%) on the MultiHop-RAG dataset. These results empirically validate the complementary contribution of fine-grained graph retrieval and coarse-grained community reports, demonstrating that the dual-channel fusion strategy is a key driver for performance enhancement in multi-hop knowledge discovery.

\section{Conclusion}

By enhancing GraphRAG with ontology-guided extraction,multi-dimensional clustering, and dual-channel fusion. OMD-GraphRAG innovatively addresses the critical challenges inherent in traditional GraphRAG frameworks:insufficient domain adaptability in cross-industry applications, poor coverage of entity relationships, and the persistent trade-off between retrieval efficiency and accuracy. Experimental results on the MultiHop-RAG benchmark demonstrate that our OMD-GraphRAG framework outperforms open-source GraphRAG frameworks by \textbf{9.21\%} in retrieval accuracy, validating the effectiveness of our method and providing the industry with a new, higher-performance open-source GraphRAG implementation.


\section{Limitations}
\paragraph{Expert-Dependent Schema Definition:} The definition of industry ontologies and schemas relies heavily on manual expertise, leading to high human resource costs and limited scalability for rapid deployment.

\paragraph{Impact of Chunking Strategies:} Since extraction is performed on pre-defined chunks, the semantic fragmentation inherent in traditional segmentation can compromise the integrity of knowledge discovery.

\paragraph{Evaluation Scope:} Results are demonstrated on a single benchmark (MultiHop-RAG); generalization to domain-specific corpora such as medical or legal QA datasets requires further validation.

\section{Ethics Statement}
This work does not involve human subjects, private user data, or any personally identifiable information. All datasets and resources used in our experiments are publicly available, license-free open-source data intended for research use. We did not collect, scrape, or process any proprietary, confidential, or restricted-access data. The proposed system is designed for retrieval-augmented question answering, and all experimental evaluations are conducted on open benchmark data. During the writing process, AI-based tools were used only for grammar checking and language polishing. They were not used to generate scientific ideas, design experiments, produce experimental results, or write substantive technical content. The authors take full responsibility for the content, methodology, analysis, and conclusions presented in this paper.


\bibliographystyle{acl_natbib}
\bibliography{custom}

\appendix
\section{Implementation Details}
\label{app:implementation}
All experiments use \texttt{Qwen3-235B} as both the extraction and generation backbone. Documents are segmented into chunks of 2{,}500 tokens with no overlap prior to graph construction. Embeddings are generated via \texttt{qwen3-embed-0.6b} and candidate reranking is performed by \texttt{qwen3-reranker-8b}, with retrieval top-$k$ set to 8 for all ablation experiments. All baselines share the same embedding model and reranker to ensure a controlled comparison; differences across systems arise solely from retrieval architecture.

\section{MultiHop-RAG Query Type Definitions}
\label{app:query_types}

MultiHop-RAG~\citep{17} is a QA dataset to evaluate retrieval and reasoning across documents with metadata in the RAG pipelines. It contains 2{,}556 queries, with evidence for each query distributed across 2 to 4 documents. The queries also involve document metadata, reflecting complex scenarios commonly found in real-world RAG applications. The dataset is released under the Open Data Commons Attribution License (ODC-BY).

\paragraph{Inference Query:} These queries are formulated by synthesizing the various characterizations of the bridge-entity across multiple claims, with the final answer being the identification of the entity itself.

\paragraph{Comparison Query:} These queries are structured to compare the similarities and differences related to the bridge entity or topic. The resultant answer to such queries is typically a definitive ``yes'' or ``no'', based on the comparison.

\paragraph{Temporal Query:} These queries explore the temporal ordering of events across different points in time. The answer to such queries is typically a ``yes'' or ``no'' or a single temporal indicator word like ``before'' or ``after''.

\paragraph{Null Query.} These queries represent cases in which the answer cannot be derived from the retrieved set. They are based on entities that do not exist in the existing bridge-entities, and fictional news source metadata is incorporated to add complexity, ensuring that the questions do not reference any contextually relevant content from the knowledge base. The answer to such queries should be ``insufficient information'' or similar.

\end{document}